\documentclass[letterpaper]{article} % DO NOT CHANGE THIS
\usepackage{aaai2027}  % DO NOT CHANGE THIS
\usepackage[hyphens]{url}  % DO NOT CHANGE THIS
\usepackage{graphicx} % DO NOT CHANGE THIS
\usepackage{natbib}  % DO NOT CHANGE THIS AND DO NOT ADD ANY OPTIONS TO IT
\usepackage{caption} % DO NOT CHANGE THIS AND DO NOT ADD ANY OPTIONS TO IT
\usepackage{algorithm}
\usepackage{algorithmic}
\usepackage{amsmath,amssymb}
\usepackage{newfloat}
\usepackage{array}
\usepackage{tabularx}
\usepackage{listings}
\DeclareCaptionStyle{ruled}{labelfont=normalfont,labelsep=colon,strut=off} % DO NOT CHANGE THIS
\floatstyle{ruled}
\newfloat{listing}{tb}{lst}{}
\floatname{listing}{Listing}

\nocopyright

\usepackage{booktabs}

\title{LEX-EC: A Lexical Evidence-Channel Audit Framework for Zero-Shot LLM Personality Classification in Black-Box Settings}
\author {
    Brittany Harbison\textsuperscript{\rm 1}\corresponding,
    Ashok K. Goel\textsuperscript{\rm 1}
}
\affiliations {
    \textsuperscript{\rm 1}Georgia Institute of
Technology \\
Atlanta, Georgia, USA\\
    bharbison3@gatech.edu, ashok.goel@cc.gatech.edu
}

\begin{document}

\maketitle

\begin{abstract}
Large language models may easily assign personality labels from text, but model interpretability remains an open problem. To address this gap, we introduce LEX-EC, a reusable black-box audit framework combining agreement diagnostics with controlled lexical ablation to distinguish marginal-distribution effects from trait-associated signal recoverable under restricted evidence. Using this framework, we illustrate how various text genres may exhibit sharply different profiles: free-form essay text contains the broadest, but still weak, signal; in graduate student introductions, an observable Extraversion association weakened after masking; and single Facebook statuses yield little stable evidence even in a trait-balanced sample, indicating a possible lower bound of content or length. Masking topical and demographic content weakened some associations while leaving others detectable from function words, affective terms, and cognitive-style vocabulary. Linguistic prompting shifted model self-explanations but did not eliminate topical content. LEX-EC jointly evaluates item-level association and chance-corrected agreement, examines how both change under lexical restriction, and conducts a targeted audit of prompt sensitivity in model-generated explanations. Across datasets, models, and prompts, LEX-EC characterizes how trait associations may vary with available lexical evidence, introducing a novel application of lexical methods to black-box interpretability in personality labeling.

\end{abstract}


\begin{links}
    \link{Code}{https://github.com/Noeru-H/LEX-EC}
\end{links}

\section{Introduction}

Large language models (LLMs) may assign personality traits \cite{goldberg1990bigfive,mccraeJohn1992ffm} to an author from ordinary text, and such inferences are increasingly used in education, hiring, and social computing \cite{vinciarelliMohammadi2014personalityComputing,matzEtAl2017psychologicalTargeting}. But a plausible label is not necessarily a valid one: a model may return the same confident judgment regardless of whether the text is rich or a single sentence, whether it reflects trait-relevant language or mere topical content, whether the inference tracks the author or demographic priors the model brings to it \cite{raoEtAl2023chatgptPersonalities,petersMatz2024psychologicalDispositions}, and whether aggregate scores mask near-chance item-level agreement. Prior work reports apparent success, as correlations with human raters or self-reports across traits and genres \cite{dernerEtAl2024chatgptRead, petersMatz2024psychologicalDispositions, schoeneggerEtAl2025aiPersonalityCorrelations, piastraCatellani2025emergentCapabilities,wrightEtAl2026assessingPersonality}, but the evidential basis of these black-box predictions remains difficult to interpret: do observable associations exist across text genres and quantities, do predictions persist under lexical content masking, and what effect does prompt engineering have on behavior? For closed or privately hosted models standard interpretability studies are impossible: without access to weights, gradients, or activations, interpretability methods that require internal inspection \cite{belinkov2022probing,olahEtAl2020circuits} do not apply. Black-box interpretability operates under this constraint, using input perturbation, attribution, and behavioral testing to characterize models from inputs and outputs alone \cite{ribeiro_2016,lundberg_lee_2017,ribeiro_2020}. This study asks what zero-shot LLM Big Five Inventory (BFI) predictions rest on once they are examined across text conditions, prompt framings, distributional checks, and content-reduced inputs. This work is organized around three questions:
\begin{itemize}
    \item \textbf{RQ1. Do zero-shot LLMs show BFI signal, and how does it vary with text length and genre?}
    \item \textbf{RQ2. When topical and demographic content is masked, which predictions collapse and which persist as recoverable from the preserved lexical layer?}
    \item \textbf{RQ3. Does linguistic prompting shift self-explanations toward stylistic and affective rationales?}
\end{itemize}
We approach these with LEX-EC (Lexical Evidence Channel auditing), a novel, reusable behavioral auditing framework for interpretability of LLMs at the task of personality prediction, applied across a text-length gradient, combining distributional and item-level agreement analysis, prompt comparison, and a content-masking ablation that tests signal survival after removal of topical and demographic vocabulary. Rather than treating LLM personality labels as measurements, they are treated solely as behavioral outputs. This framework can narrow causal hypotheses, motivate targeted follow-up interventions, and provide evidence informing decisions about under what conditions personality-labeling systems should be used.

\section{Related Work}

\subsection{Personality Measures and the Big Five Inventory}
BFI models personality along Extraversion, Neuroticism, Agreeableness, Conscientiousness, and Openness through questionnaire responses \cite{johnEtAl1991bfi,mccrae_costa_1987,john_srivastava_1999}. Although these traits are ordinarily measured continuously, some reference datasets used in personality labeling historically provide only binary high/low labels. Prior psycholinguistic work has found modest, context-dependent associations between personality and function words, affective vocabulary, and other linguistic patterns \cite{pennebaker_king_1999,yarkoni_2010,schwartz_2013}, motivating the possibility of personality prediction from text.

\subsection{Lexicon-based Textual Analysis: LIWC, Emotion, NRC}
Lexicon-based methods such as LIWC map words to psychologically motivated categories including affect, cognition, and function-word use \cite{tausczik_pennebaker_2010}. Open resources such as Empath and NRC-EmoLex provide related affective and cognitive categories \cite{fast_2016,mohammad_turney_2013}. Notably, Empath's categories, when compared against analogous LIWC categories, showed a reported average correlation of $r=.906$ \cite{fast_2016}. We use these resources to construct the preserved lexical layer in our masking ablation.

Corpus-linguistic studies show that text length is systematically
associated with register, linguistic-feature distributions, and
communicative function rather than varying independently of the
setting in which text is produced
\cite{liimatta2022registers,liimatta2023variation,
ohmanliimatta2024intentionality}, rendering these characteristics difficult to separate without confounding.

\subsection{LLM Personality Prediction}
Recent work suggests that LLMs can recover nonzero BFI signal from text, sometimes approaching individual human judgments \cite{dernerEtAl2024chatgptRead,schoeneggerEtAl2025aiPersonalityCorrelations, piastraCatellani2025emergentCapabilities,wrightEtAl2026assessingPersonality}. Prompting strategies can materially affect this performance \cite{yang-etal-2023-psycot}. Reported performance varies substantially with prompt framing, text type, text quantity, and the evaluation target \cite{bhandarkarEtAl2025psytex, piastraCatellani2025emergentCapabilities}. Models also exhibit demographic sensitivity, positivity bias, poor confidence calibration, and weaker agreement with self-reported traits than with individual human ratings \cite{raoEtAl2023chatgptPersonalities,petersMatz2024psychologicalDispositions}. In particular, prior studies commonly aggregate multiple messages or extended texts per author, leaving performance under limited-text conditions less well characterized;linguistic and affective evidence from topical or demographic cues are also rarely examined. We therefore evaluate predictions across a text-length gradient and test whether apparent signal persists after content masking.

Complementing prompt-based studies, Maharjan et al.\ trained classifiers over BERT, RoBERTa, and OpenAI embeddings on the PANDORA dataset, reporting stronger performance than zero-shot prompting and associations with LIWC and NRC features, although psychometric reliability was only moderate \cite{maharjanEtAl2025psychometric}.

\begin{figure*}[!t]
    \centering
    \includegraphics[width=\textwidth]{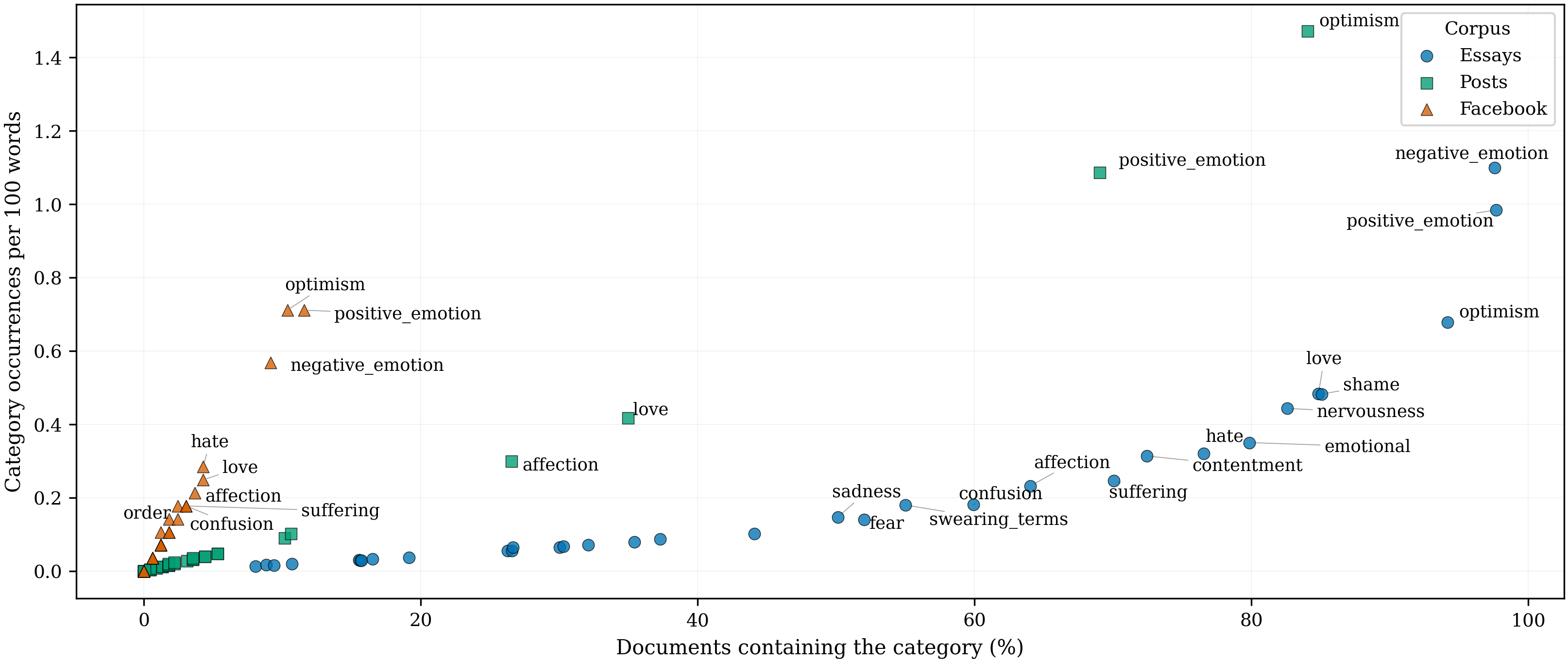}
    \caption{Document prevalence and corpus-level density of retained Empath and
NRC categories. Prevalence is the percentage of documents containing at least
one category term; density is the number of category occurrences per 100 word-tokens. Lexicon categories are not mutually exclusive. Labels identify selected
high-prevalence, high-density, or corpus-distinctive categories.}
    \label{fig:dataset_analysis}
\end{figure*}

\subsection{Black-box Interpretability}
Because closed LLMs do not expose weights or activations, behavior is often characterized through controlled input transformations. Common black-box
interpretability methods include local feature-attribution approaches such
as LIME and SHAP, which estimate the contributions of input features to
individual predictions
\cite{ribeiro_2016,lundberg_lee_2017}, as well as behavioral-testing
approaches such as CheckList, which evaluate model responses under targeted
input conditions and transformations \cite{ribeiro_2020}. Input-reduction
studies similarly examine whether predictions persist as information is
removed, while showing that behavior on transformed inputs need not provide
a faithful account of the process that produced the original prediction
\cite{fengEtAl2018pathologies}. Some work additionally argues that interpretability methods should be evaluated in relation to the particular claims and contexts they are
intended to support \cite{doshi-velez2018considerations}.

Model-generated explanations present a related, but distinct, black-box
signal. Fluent self-explanations may not faithfully reflect the computation
producing a prediction
\cite{jacoviGoldberg2020faithfulness,turpinEtAl2023unfaithfulCot}, though they may still reflect some signal in aggregate \cite{madsen-etal-2024-self}.

Comparative evidence indicates that lexicon-based measures are useful in NLP because they provide transparent, computationally inexpensive, and directly inspectable representations of language \cite{vanDerVeenBleich2025advantages}. Related work in quantitative discourse analysis likewise emphasizes the value of explicit lexical evidence for improving methodological transparency and making analytical decisions easier to trace \cite{compton2025beyond}. Relatedly, PsyTEx removes spans chosen from LLM-generated trait criteria and repeats prediction on the remainder, subsequently analyzing the selected text with LIWC \cite{bhandarkarEtAl2025psytex}. LEX-EC instead specifies the retained psycholinguistic channel independently to avoid self-report dependence and couples lexical restriction with item-level agreement audits.

\section{Datasets}
Three datasets were selected to span distinct genres and average text lengths: single MyPersonality Facebook statuses \cite{celliEtAl2013sharedTask}, graduate student forum introductions \cite{X2026personality}, and Pennebaker-King essays \cite{pennebaker_king_1999, celliEtAl2013sharedTask}. The texts average 17, 89, and 672 mean retained word-tokens \footnote{Word-tokens: spaCy tokens on preprocessed text (URLs stripped, PII and program identifiers replaced), excluding punctuation, whitespace, and emoticon/emoji spans.}, respectively, representing micro-, short-, and middle-length conditions. This places the study at the lower end of the text quantities used in prior zero-shot personality-prediction work, which often aggregates multiple letters, statuses, or other texts per author. Binary high/low labels provide a common prediction and evaluation target across corpora.

The corpora also differ in retained psycholinguistic content (Figure~\ref{fig:dataset_analysis}). Essays show broader document-level coverage of several affective categories; Facebook statuses show lower prevalence but sometimes substantial density when a category occurs; and student introductions contain comparatively frequent optimism-related language. Thus, the conditions vary in lexical composition as well as length and genre.

\subsection{Student Introduction Posts}
We collected BFI-44 responses \cite{johnEtAl1991bfi} from students across two semesters of Georgia Tech’s OMSCS KBAI course. Of these, 226 also posted course-forum introductions, yielding complete paired survey–post records. Participation was voluntary and consented, procedures were IRB-approved, and names, emails, and links were removed before analysis. BFI-44 scores were converted to binary high/low labels using the midpoint procedure from prior work \cite{X2026personality}, which is included in the Supplementary material.

The cohort’s trait distributions were uneven, with marked right skew for all traits except Extraversion and Neuroticism; only Extraversion, Neuroticism, and Conscientiousness were sufficiently distributed for trait-level analysis. The introduction posts are a distinctive self-presentational genre: public educational forums encourage impression management \cite{hayesEtAl2024selfPresentation}, and the icebreaker template solicited demographic and biographical details including location, courses, specialization, hobbies, motivation, and an interesting fact.

\begin{figure*}[!t]
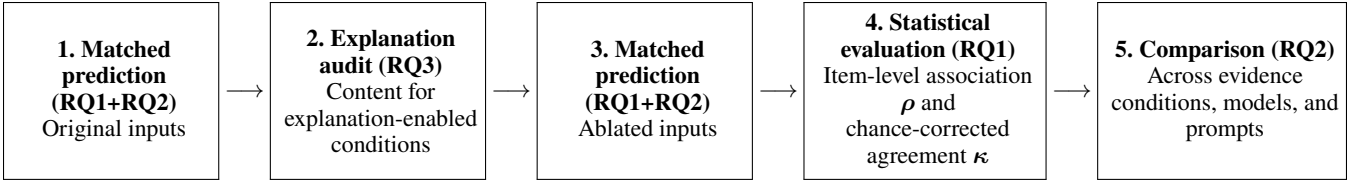

\centering
\setlength{\fboxsep}{2.5pt}
\setlength{\tabcolsep}{0pt}

{\small
\begin{tabular*}{\textwidth}{
    @{\extracolsep{\fill}}
    c c c c c c c c c
    @{}
}

\fbox{
\parbox[c][2.15cm][c]{0.145\textwidth}{
\centering
\textbf{1. Matched prediction (RQ1+RQ2)}\\
Original inputs
}}
&
$\longrightarrow$
&
\fbox{
\parbox[c][2.15cm][c]{0.145\textwidth}{
\centering
\textbf{2. Explanation audit (RQ3)}\\
Content for explanation-enabled conditions
}}
&
$\longrightarrow$
&
\fbox{
\parbox[c][2.15cm][c]{0.145\textwidth}{
\centering
\textbf{3. Matched prediction  (RQ1+RQ2)}\\
Ablated inputs
}}
&
$\longrightarrow$
&
\fbox{
\parbox[c][2.15cm][c]{0.165\textwidth}{
\centering
\textbf{4. Statistical evaluation (RQ1)}\\
Item-level association $\boldsymbol{\rho}$ and chance-corrected agreement
$\boldsymbol{\kappa}$}}
&
$\longrightarrow$
&
\fbox{
\parbox[c][2.15cm][c]{0.17\textwidth}{
\centering
\textbf{5. Comparison (RQ2)}\\
Across evidence conditions, models, and prompts
}}

\end{tabular*}
}

\caption{LEX-EC processing stages. Statistical analyses address RQ1, full-versus-masked comparisons address RQ2,
and self-explanation analysis addresses RQ3.}
\label{fig:lexec-stages}
\end{figure*}

\subsection{Free-form Essays}
The essay corpus is the Pennebaker–King stream-of-consciousness dataset \cite{pennebaker_king_1999}: open-ended writing produced under a fixed-period, write-whatever-comes-to-mind protocol, paired with BFI labels. It serves as our long-form condition, averaging ~672 word-tokens (\(\approx 3{,}295\) characters) per text after preprocessing.

We report essay results at two scales. A 226-item subset, drawn by proportional stratification over joint five-trait label profiles with a fixed seed, preserves the full corpus's trait-profile distribution and is used for pilot prompt and model comparisons; unlike the Facebook sample, it is a proportional downsample. The full corpus (n=2468) provides the more stable estimate and supersedes the pilot subset for all trait-level essay claims; the subset serves only to check that pilot behavior is consistent with sampling variability from the full set.

\subsection{Facebook Statuses}
The Facebook corpus derives from the myPersonality project
\cite{celliEtAl2013sharedTask}, which linked Facebook status updates to user-level BFI
questionnaire labels. After sampling one status per user with fixed seed $67$,
the corpus provides a naturally occurring microtext condition averaging
approximately 17 word-tokens (91 characters) per text after preprocessing.

We simultaneously balanced the five marginal trait distributions using
mixed-integer linear programming. Let $\mathcal{P}$ be the set of observed
five-trait binary profiles, $n_p$ the number of users with profile $p$, and
$a_{jp}\in\{0,1\}$ indicate whether profile $p$ has a positive label for trait
$j$. We selected the number $z_p$ of users retained from each profile by
solving
\begin{equation}
\begin{aligned}
\max_{\mathbf{z}}\quad
    & N=\sum_{p\in\mathcal{P}} z_p \\[2pt]
\text{s.t.}\quad
    & z_p\in\{0,\ldots,n_p\},
      && p\in\mathcal{P},\\
    & (0.5-\delta)N
      \leq \sum_{p\in\mathcal{P}} a_{jp}z_p,
      && j\in\mathcal{T},\\
    & \sum_{p\in\mathcal{P}} a_{jp}z_p
      \leq (0.5+\delta)N,
      && j\in\mathcal{T},
\end{aligned}
\end{equation}
where $\mathcal{T}$ is the set of five traits and $\delta=0.05$. With no
fixed target size, the solution attained the largest feasible sample. We solved the program with \texttt{scipy.optimize.milp} using its HiGHS
backend, sampled users
without replacement within each profile, and used fixed seed $67$ for all sampling. The resulting dataset contains 164
unique users, with each trait's positive-label proportion between 45\% and
55\%.

\section{Methodology}
Figure~\ref{fig:lexec-stages} summarizes the LEX-EC workflow.
Across the three datasets, models assign binary BFI labels to
original and content-masked texts under matched prompt conditions.
We separately inspect item-level association and chance-corrected agreement, full-to-masked changes, and explanation content where collected. Each observation consists of one text from one author, and
no model is fine-tuned on the target datasets. Model outputs in both conditions were collected through required tool calls.\footnote{Strict schema-constrained decoding in the API was avoided because
it degraded association statistics in pilot tests.} Unless otherwise stated, all results were each gathered via a single run.

\subsection{Model Selection}
Proprietary LLM evaluation presents a reproducibility problem: model families
are revised frequently, older endpoints may be deprecated, and implementation
details are generally unavailable. A result obtained from one model snapshot
may therefore reflect transient provider- or version-specific behavior rather
than a stable property of the task. We address this major reproducibility issue in the field as it applies to our work by evaluating
five closed models spanning providers, model generations, and prediction
paradigms: GPT-4o-mini \cite{openai2024gpt4omini}, GPT-4o \cite{openai2024gpt4osystemcard}, o3-mini \cite{openai2025o3minisystemcard}, GPT-5.4-mini \cite{openai2026gpt54systemcard}, and Claude Haiku 4.5 \cite{anthropic2025haiku45systemcard}.\footnote{Snapshots used: \textit{gpt-4o-mini-2024-07-18}, \textit{gpt-4o-2024-11-20}, \textit{o3-mini-2025-01-31}, \textit{gpt-5.4-mini-2026-03-17}, \textit{claude-haiku-4-5-20251001}.} \footnote{Models queried with $temperature=0$ for non-reasoning models, however this was not possible for reasoning models. $max\_tokens=3000$, $max\_completion\_tokens=3000$ used for reasoning models. Reasoning models were set to an effort of $medium$. All used the default $top_p=1$. No fixed seed was supplied.}

GPT-4o-mini and GPT-4o provide compact and larger-model baselines from the
same pre-reasoning family; o3-mini and GPT-5.4-mini provide
reasoning-oriented and newer-generation contrasts; and Claude Haiku provides
a cross-provider comparison. The purpose is to test whether the main evidence-channel findings
may persist across model snapshots that differ in scale, provider, and generation.

Because the pilot comparisons showed similar dataset-level patterns, with limited descriptive variation in the direction and magnitude of Spearman’s $r$ and Cohen’s $\kappa$ across models, the full-corpus essay analysis was run with GPT-5.4-mini under the linguistic prompt rather than repeated for every model.

\subsection{Prompts}
Prompts directed the model to adopt an expert persona for this task; prompting was treated as a sensitivity factor rather than optimized for a single “best” formulation. We compared \textit{Basic} (BASIC), requesting binary BFI classifications; \textit{Basic+Explanation} (BASICEX), adding trait-level justifications; and \textit{Linguistic+Explanation} (LINEX), directing justifications toward linguistic, stylistic, and affective rather than topical or demographic evidence. Although findings on the efficacy of
expert personas are mixed, persona prompting can alter model output
behavior, with effects that vary by use
\cite{zheng-etal-2024-helpful,kong-etal-2024-better,
luz-de-araujo-etal-2025-principled}.
BASICEX therefore matched LINEX in its
expert-role and justification instructions, differing only in the
instruction to prioritize linguistic evidence. 

A preliminary linguistic-only condition was dropped after showing no descriptive performance difference from its explanation-bearing counterpart, as was BASICEX, though it remained in use only as the explanation-bearing baseline for the targeted RQ3 self-explanation audit.

\subsection{Model Self-Explanations}
For explanation-enabled conditions, GPT-4o-mini provided a brief justification alongside each binary label. We analyzed explanations only for Extraversion in the student-introduction dataset, the sole post-level trait with an interpretable predictive signal; label imbalance or lack of trait discrimination made the remaining traits unsuitable for explanation analysis. This is therefore a targeted audit of one trait rather than a five-trait analysis.

We manually coded, with a single annotator, GPT-4o-mini
Extraversion explanations for the student introduction posts
as linguistic or linguistic-adjacent (LIN), demographic or
topical (DEM), combined (CMB), or other (OTH).

We treat self-explanations as aggregate behavioral outputs only, as post hoc LLM explanations may be unfaithful \cite{jacoviGoldberg2020faithfulness,turpinEtAl2023unfaithfulCot,madsen-etal-2024-self}. We test whether the distribution of stated self-explanation types, such as linguistic or affective versus demographic or topical, shifts across prompt conditions. A shift toward linguistic justifications under the linguistic prompt indicates a change in explanation behavior, providing evidence of prompt sensitivity.

\subsection{Performance and Association Metrics}
Because the reference labels may be imbalanced in real-world data, we assess predictions
using complementary measures of item-level association and chance-corrected
agreement. Our primary measures are
\begin{equation}
\rho_s
=
\operatorname{corr}\!\left(
\operatorname{rank}(\mathbf{y}),
\operatorname{rank}(\hat{\mathbf{y}})
\right),
\qquad
\kappa
=
\frac{p_o-p_e}{1-p_e},
\end{equation}
where $\mathbf{y}$ and $\hat{\mathbf{y}}$ are the reference and predicted
labels, respectively, $p_o$ is observed agreement, and $p_e$ is agreement
expected from the observed marginals. For binary variables, Spearman's
$\rho$ is equivalent to the phi coefficient, while Cohen's $\kappa$ measures
agreement after accounting for marginal prevalence.

We test dependence between predicted and reference labels using Fisher's
exact test for sparse $2\times2$ tables, and Pearson's $\chi^2$ test otherwise. We interpret $p$-values as evidence against independence rather
than as measures of practical strength, which is assessed using the
magnitudes of $\rho$ and $\kappa$.

\subsection{LLM Personality Trait Prediction}
The first prediction stage used the unablated text: for each dataset, prompt condition, and model, the LLM produced binary BFI labels from the full text available in that condition, establishing pre-ablation performance, per Algorithm~\ref{alg:prediction_tool}.

\begin{algorithm}[!t]
\caption{Zero-shot BFI prediction}
\label{alg:prediction_tool}
\begin{algorithmic}[1] \small
\REQUIRE Texts $X=\{x_i\}$, model $m$, prompt condition $c$
\ENSURE Predictions $\hat{y}_{i,t}\in\{0,1\}$ and optional explanations $j_{i,t}$
\STATE $T \leftarrow \{\mathrm{O},\mathrm{C},\mathrm{E},\mathrm{A},\mathrm{N}\}$
\STATE Select prompts $(s_c,u_c)$ and format example $F_c$
\STATE Define required tool schema $S_c$ for all $t\in T$, including
       \texttt{justification} when required by $c$
\FOR{each text $x_i\in X$}
    \STATE $q_i \leftarrow u_c \mathbin{+\!\!+} F_c
           \mathbin{+\!\!+} x_i$
    \STATE $a_i \leftarrow
           \mathrm{ToolCall}(m,s_c,q_i,S_c)$
    \STATE $z_i \leftarrow \mathrm{Parse}_m(a_i)$
    \FOR{each trait $t\in T$}
        \STATE $\hat{y}_{i,t}
               \leftarrow
               \mathbb{I}\!\left[
               \mathrm{low}(z_i[t].\texttt{classification})
               =\texttt{high}\right]$
        \IF{$c$ requires explanations}
            \STATE $j_{i,t}\leftarrow z_i[t].\texttt{justification}$
        \ENDIF
    \ENDFOR
    \STATE Store predictions, optional explanations, model, prompt,
           dataset, and text identifier
\ENDFOR
\RETURN $\{(\hat{y}_{i,t},j_{i,t})\}$
\end{algorithmic}
\end{algorithm}

\begin{table*}[!t]
\centering
\small

{
\setlength{\tabcolsep}{3pt}

\begin{tabular*}{\textwidth}{
    @{\extracolsep{\fill}} l c c c c c @{}
}
\hline\hline
Metric & O & C & E & A & N \\
\hline

\multicolumn{1}{l}{} &
\multicolumn{5}{l}{\hspace{0.5em}\textit{Essays, full corpus: GPT-5.4-mini, LINEX}} \\

$\rho$
& $.168{\to}.158\;[-.010]$
& $.220{\to}.121\;[-.099]$
& $.187{\to}.147\;[-.040]$
& $.151{\to}.150\;[-.001]$
& $.115{\to}.083\;[-.032]$ \\

$\kappa$
& $.124{\to}.145\;[+.021]$
& $.220{\to}.119\;[-.101]$
& $.187{\to}.144\;[-.043]$
& $.135{\to}.136\;[+.001]$
& $.044{\to}.026\;[-.018]$ \\

\hline
\multicolumn{1}{l}{} &
\multicolumn{5}{l}{\hspace{0.5em}\textit{Facebook statuses: GPT-4o, BASIC ($n=164$)}} \\

$\rho$
& $-.020{\to}.066\;[+.086]$
& $.163{\to}.088\;[-.075]$
& $.064{\to}.028\;[-.036]$
& $-.016{\to}-.027\;[-.011]$
& $.184{\to}-.058\;[-.242]$ \\

$\kappa$
& $-.020{\to}.052\;[+.072]$
& $.148{\to}.044\;[-.104]$
& $.063{\to}.025\;[-.038]$
& $-.016{\to}-.022\;[-.006]$
& $.183{\to}-.056\;[-.239]$ \\

\hline
\multicolumn{1}{l}{} &
\multicolumn{5}{l}{\hspace{0.5em}\textit{Posts: GPT-4o-mini, LINEX ($n=226$)}} \\

$\rho$
& $-.045{\to}-.035\;[+.010]$
& $-.026{\to}-.111\;[-.085]$
& $.226{\to}.062\;[-.164]$
& $-.036{\to}.045\;[+.081]$
& $.024{\to}.078\;[+.054]$ \\

$\kappa$
& $-.031{\to}-.032\;[-.001]$
& $-.023{\to}-.068\;[-.045]$
& $.173{\to}.058\;[-.115]$
& $-.017{\to}.044\;[+.061]$
& $.009{\to}.054\;[+.045]$ \\

\hline\hline
\end{tabular*}
}

\caption{Primary association and agreement results before and after content
ablation. Cells report Full $\rightarrow$ Ablated $[\Delta]$, where
$\Delta=\mathrm{Ablated}-\mathrm{Full}$. O=Openness,
C=Conscientiousness, E=Extraversion, A=Agreeableness, and N=Neuroticism. Shown are: full set essays with GPT-5.4-mini/LINEX after similar pilot results across models; posts with GPT-4o-mini/LINEX; Facebook with GPT-4o/BASIC, which showed the clearest pre-ablation signal and its subsequent attenuation. Deltas use coefficients rounded to three decimal places.}
\label{tab:primary-ablation}
\end{table*}

\begin{table*}[!t]
\centering
\footnotesize
\setlength{\tabcolsep}{4pt}
\small

\begin{tabular*}{\textwidth}{
@{\extracolsep{\fill}} l c c c c c @{}
}
\hline\hline
Dataset (pairs) & O & C & E & A & N \\
\hline

Posts (10)
& $0{\to}1;\;+.011/-.001$
& $0{\to}0;\;-.071/-.055$
& $7{\to}2;\;-.078/-.093$
& $0{\to}0;\;+.081/-.005$
& $0{\to}0;\;+.054/+.022$ \\

Essays (10)
& $10{\to}9;\;-.024/+.004$
& $4{\to}1;\;-.044/-.053$
& $6{\to}3;\;-.034/-.039$
& $8{\to}7;\;+.014/+.008$
& $6{\to}4;\;-.043/-.049$ \\

Facebook (2)
& $0{\to}0;\;+.069/+.052$
& $1{\to}0;\;-.044/-.074$
& $0{\to}0;\;+.006/.000$
& $0{\to}0;\;-.022/-.020$
& $2{\to}0;\;-.173/-.171$ \\

\hline\hline
\end{tabular*}

\caption{Cross-model summary of changes after content ablation.
Cells report significant Spearman associations,
Full$\to$Ablated, followed by median
$\Delta\rho/\Delta\kappa$ across matched model--prompt pairs.
Posts and reduced-set essays include five models under two prompts;
Facebook includes GPT-4o under two prompts.
$\Delta=\mathrm{Ablated}-\mathrm{Full}$. Threshold for significance was considered to be $\rho \approx .1307$ for Essays/Posts, $\approx .153$ for Facebook. Deltas use coefficients rounded to three decimal places; pairs with undefined $\rho$ were omitted from median $\Delta\rho$.
}
\label{tab:crossmodel-ablation-summary}
\end{table*}

\subsection{Content-Masking Ablation}
The preserved lexical layer was constructed from selected Empath
Emotion Lexicon categories, selected to closely follow LIWC and the noted strong correlation between them.
\cite{fast_2016,
tausczik_pennebaker_2010}. The channel therefore draws on validated lexical
categories and an established psycholinguistic framework, granting it a stronger
a priori construct-validity basis than an ad hoc or generic bag-of-words
method. NRC categories \cite{mohammad_turney_2013} were additionally selected and added to the layer to address lexical coverage gaps of specific Emotion categories not granted by the Empath library.

The retained psycholinguistic vocabulary is represented by,

\begin{equation}
V_{\mathrm{psych}}
=
\bigcup_{c \in C_E} \ell_E(c)
\;\cup\;
\left\{
w \in W_{\mathrm{NRC}} :
\ell_N(w) \cap K \neq \varnothing
\right\},
\label{eq:retained-vocab}
\end{equation}

where $C_E$ contains the selected Empath affective and cognitive-style
categories, $K$ the retained NRC emotion and polarity categories,
$\ell_E(c)$ the Empath word set for category $c$, and $\ell_N(w)$ the NRC tags
assigned to $w$.

Each text is transformed by,

\begin{equation}
f
=
\Lambda
\circ \sigma
\circ \operatorname{Recon}_{\tau}
\circ \operatorname{nlp}
\circ \rho_{\mathrm{prog}}
\circ \rho_{\mathrm{pii}}
\circ \rho_{\mathrm{tmpl}}
\circ \rho_{\mathrm{url}},
\label{eq:masking-pipeline}
\end{equation}

where $\rho_{\mathrm{url}}$ removes URLs;
$\rho_{\mathrm{tmpl}}$ removes copied icebreaker questions after normalizing
punctuation variants; $\rho_{\mathrm{pii}}$ replaces PII placeholders; and
$\rho_{\mathrm{prog}}$ masks program and course identifiers, degree
abbreviations, and course numbers with placeholder $\mu$. Template removal
applies only to student introductions, preventing copied prompts from being
treated as authored evidence.

After tokenization and POS tagging, tokens are mapped top-down by,

\begin{equation}
\tau(t)
=
\begin{cases}
t,
& E(t) \lor Q(t), \\

t,
& \operatorname{pos}(t) \in \Pi_F
  \lor \operatorname{low}(t) \in W_F, \\

t,
& \left\{
    \operatorname{low}(t),
    \operatorname{low}(\operatorname{lem}(t))
  \right\}
  \cap V_{\mathrm{psych}}
  \neq \varnothing, \\

\mu,
& \operatorname{pos}(t) = \mathrm{NUM}, \\

\mu,
& \text{otherwise}.
\end{cases}
\label{eq:token-mapping}
\end{equation}

Here, $E(t)$ and $Q(t)$ identify emoji or emoticons and punctuation or
whitespace; $\Pi_F$ contains retained function-word POS classes; and $W_F$
contains supplemental quantifiers, negations, degree and temporal adverbs, and
wh-words. Numerals are masked because they may encode ages, years, or counts.
$\operatorname{Recon}_{\tau}$ restores original whitespace, $\sigma$ collapses
consecutive placeholders, and $\Lambda$ removes empty or placeholder-only
lines. The masking placeholder $\mu$ was instantiated as the literal
underscore character '\texttt{\_}' to avoid introducing lexical associations that may accompany replacement with unrelated words, or adjacency issues arising from simple word deletion.

The resulting ablation is a retention test: all material outside the
function-word, affective, cognitive-style, and structural channel is masked.
Persistence indicates that an association remains recoverable from the
retained channel; attenuation indicates that broader lexical content
contributed to the unmasked association.

\section{Experiments}

\subsection{Model Self-Explanations}
Under the basic explanation
prompt, 113 of 226 explanations (50.0\%) were coded LIN; under the linguistic
prompt, this increased to 153 of 226 (67.7\%). Explanations containing any
demographic or topical reasoning (DEM or CMB) decreased from 113 of 226
(50.0\%) to 72 of 226 (31.9\%). Linguistic prompting shifted the
content of the generated explanations without eliminating topical reasoning.

Explanation type did not reliably distinguish correct from incorrect
predictions: linguistic and demographic/topical justifications appeared in both
groups. Linguistic prompting showed the same pattern. For example, sports and
fitness references appeared equally often in correct-high and incorrect-high
explanations under the basic prompt (14 each), and appeared across multiple
outcome categories under the linguistic prompt.

Prediction correlations also showed prompt sensitivity, but changes between the BASICEX and LINEX conditions varied across models, with no consistent advantage for either prompt.

\subsection{Prediction and Ablation Results}
Prediction patterns differed substantially across datasets
(Tables~\ref{tab:primary-ablation} and
\ref{tab:crossmodel-ablation-summary}). In the student introduction posts,
Extraversion was the only trait with statistically detectable pre-ablation
associations in a majority of model--prompt conditions (seven of ten).
After masking, only two of ten Extraversion associations remained
detectable. Both its association and agreement coefficients decreased in
eight of ten conditions, with median paired changes of
$\Delta\rho=-.078$ and $\Delta\kappa=-.093$. In the focal
GPT-4o-mini/LINEX condition, Extraversion decreased from
$\rho=.226$ to $.062$ and from $\kappa=.173$ to $.058$.
The other traits remained inconsistent across conditions, with generally
negligible chance-corrected agreement and several constant or near-constant
prediction distributions.

Essays produced broader associations. In the reduced-set model comparison,
positive pre-ablation coefficients appeared across multiple traits, although
their magnitudes remained small. In the full-corpus GPT-5.4-mini analysis
under the linguistic prompt, the Spearman associations for all five traits
were statistically detectable both before and after masking. The largest
decrease occurred for Conscientiousness
($\rho$: $.220\rightarrow.121$;
$\kappa$: $.220\rightarrow.119$).
Extraversion also decreased
($\rho$: $.187\rightarrow.147$;
$\kappa$: $.187\rightarrow.144$), whereas Agreeableness changed little
($\rho$: $.151\rightarrow.150$;
$\kappa$: $.135\rightarrow.136$).
For Openness, the association was nearly unchanged
($\rho$: $.168\rightarrow.158$), while chance-corrected agreement increased
from $.124$ to $.145$. Neuroticism had the weakest agreement both before
and after masking
($\kappa$: $.044\rightarrow.026$).

The Facebook results were weak across most model--prompt--trait
combinations. Four of the 50 pre-ablation associations were statistically
detectable: Conscientiousness under Claude Haiku/LINEX and GPT-4o/BASIC,
and Neuroticism under GPT-4o/BASIC and GPT-4o/LINEX. Matched Facebook
ablation analyses were available for the two GPT-4o prompts; after masking,
none of their associations remained statistically detectable. In the focal
GPT-4o/BASIC condition, Conscientiousness decreased from
$\rho=.163$ and $\kappa=.148$ to $\rho=.088$ and $\kappa=.044$, while
Neuroticism changed from $\rho=.184$ and $\kappa=.183$ to negative values
($\rho=-.058$, $\kappa=-.056$). Under the linguistic prompt,
Neuroticism also decreased
($\rho$: $.166\rightarrow.062$;
$\kappa$: $.164\rightarrow.062$).
The remaining Facebook coefficients changed in both directions but were
small before and after masking.

Overall, masking most consistently attenuated Extraversion in the
introduction posts and Conscientiousness in the full-corpus essay analysis.
Openness and Agreeableness remained comparatively stable in the focal essay
condition, while the limited pre-ablation Facebook associations were not
maintained after masking.

Across models and prompts, the reduced-set essay results varied in
magnitude but showed the same broad trait-level pattern.

In the Facebook condition, 59.8\% of lexical tokens survived masking, comparable to the student posts. However, statuses were very short, the resulting inputs averaged only 10.2 retained tokens, including $\approx$ 1.9 psycholinguistic-vocabulary tokens. Post-ablation, mean retained word-token counts for Essays were 471, Posts 54, Facebook 10.

\section{Discussion}
Essays produced the broadest associations, but small chance-corrected agreement, indicating that statistical detectability did not translate into reliable individual-level classification. Student introductions yielded a narrower pattern centered on Extraversion, perhaps because this genre naturally elicits descriptions of social activities and preferences. Facebook performance remained weak despite balanced reference labels, showing a possible lower length/content boundary condition.

The ablation results further suggest that different traits are consistent with dependence on different linguistic cues. In the full-corpus essay analysis, Conscientiousness was most attenuated, consistent with greater dependence on removed content-bearing vocabulary. Agreeableness was comparatively stable, while the near-stable Openness association accompanied by increased $\kappa$ more likely reflects a change in prediction marginals than stronger trait information. In student introductions, the repeated decline in Extraversion across model--prompt conditions suggests sensitivity to comparatively explicit lexical cues. The disappearance of the isolated Facebook associations is likewise consistent with dependence on sparse lexical evidence, although the short retained inputs and limited matched ablation conditions preclude a stronger explanation.

No model showed a stable advantage across datasets and conditions; model differences were less systematic than differences associated with genre, trait, and masking. This suggests that the pattern of findings may generalize to newer model families.

Linguistic prompting increased linguistic explanations, while explanations containing demographic or topical
reasoning decreased. Prompt framing therefore changed the distribution of stated justifications without directly establishing that those justifications reflected
the evidence responsible for the classifications
\cite{jacoviGoldberg2020faithfulness,turpinEtAl2023unfaithfulCot}.

The central result is not simply that performance was weak: associations differed in their sensitivity to lexical masking. Some remained observable from the retained function-word, psycholinguistic, and structural channel, whereas others attenuated when broader lexical evidence was removed. The broadly similar patterns across models suggest that weak performance in some conditions may reflect limitations of the available textual evidence rather than a deficiency specific to one model.

These findings suggest that zero-shot personality prediction is best characterized as a property of a specific configuration of model, genre, trait, available word usage than as a general capability of a model. Persistence under lexical restriction is not inherently evidence of better personality prediction: it identifies signal recoverable from the retained channel, which may include differing stylistic contents. Personality-labeling systems should thus be evaluated separately for each intended text condition, rather than treated as possessing a general personality prediction capability.

\subsection{Limitations}
The explanation audit was manually coded by one annotator, so the
reported explanation-category frequencies and prompt-related shifts lack an independent inter-rater reliability estimate.

Exposure of closed models to the public personality corpora is unknown; results on those corpora cannot establish fully uncontaminated
out-of-sample performance. The private student-introduction corpus was immune to this effect. Additionally, by removing all but psycholinguistic material during ablation, this may disrupt correspondence with data consumed during training.

\section{Conclusion}
Zero-shot BFI prediction from a single text is not a unitary capability: evidential profiles varied across corpora and traits, and lexical restriction produced heterogeneous changes. Linguistic prompting also shifted model self-explanations toward stylistic and affective evidence, indicating prompt sensitivity in explanation behavior. 

These results demonstrate the utility of LEX-EC as a task-specific black-box audit framework. By examining agreement, lexical-channel persistence, and explanation content together, our framework revealed information invisible to conventional statistical measures alone.

\section*{AI Tool Use Disclosure}
The authors utilized GPT 5.5-5.6, Claude Opus 4.8, and Claude Fable 5 while editing text for clarity, grammar, and concision, and as non-decisional adversarial review aids to flag assumptions/methodological weaknesses in both the text and suggestions. Most non-copyediting suggestions were rejected as scientifically invalid/out-of-scope by the authors; suggestions that did inform revisions were human-verified.

\bibliography{aaai2027}

\appendix
\clearpage
\section{Appendix}
\subsection{Midpoint for Student Introduction Posts}
Before use, the student post dataset was converted from the original continuous data to binary high/low. Each personality-trait score was dichotomized using the fixed
scale midpoint of $\tau=3$. For participant $i$ and personality
trait $j$, the binary classification $c_{ij}$ was defined as,

\begin{equation}
c_{ij}
=
\mathbb{I}(x_{ij}>3)
=
\begin{cases}
0, & \text{if } x_{ij} \leq 3,\\
1, & \text{if } x_{ij} > 3,
\end{cases}
\label{eq:trait-binarization}
\end{equation}

where $x_{ij}$ denotes the original personality-trait score and
$\mathbb{I}$ denotes the indicator function. Thus, scores less
than or equal to the scale midpoint were assigned to class~0,
whereas scores above the midpoint were assigned to class~1. Past work utilizing this dataset \cite{X2026personality}, instead used a random midpoint procedure, which can introduce stochastic label noise, and was therefore avoided in this work.

\subsection{Additional Dataset Distributions}
This section includes additional tables. Table~\ref{tab:intro-post-class-distribution} demonstrates the distribution of the Student Introduction Post dataset, introduced in Section "Datasets" in the main body. Table~\ref{tab:facebook_distro} shows the distribution data for the marginally balanced single-post-per-user subset, while Table~\ref{tab:essays-class-distribution} and Table~\ref{tab:essays-subset} show the distribution data for the full essay set and for the subset, respectively.

\begin{table}[t]
\centering
\small
\begin{tabular*}{\columnwidth}{
    @{\extracolsep{\fill}}lcr@{}
}
\toprule
\textbf{Trait} & \textbf{Value} & \textbf{Proportion} \\
\midrule
\textbf{O}
    & 1 & 0.898230 (89.82\%) \\
    & 0 & 0.101770 (10.18\%) \\
\addlinespace

\textbf{C}
    & 1 & 0.840708 (84.07\%) \\
    & 0 & 0.159292 (15.93\%) \\
\addlinespace

\textbf{E}
    & 1 & 0.438053 (43.81\%) \\
    & 0 & 0.561947 (56.19\%) \\
\addlinespace

\textbf{A}
    & 1 & 0.871681 (87.17\%) \\
    & 0 & 0.128319 (12.83\%) \\
\addlinespace

\textbf{N}
    & 1 & 0.429204 (42.92\%) \\
    & 0 & 0.570796 (57.08\%) \\
\bottomrule
\end{tabular*}
\caption{Distribution of binary Big Five personality labels in the
student introduction-post dataset ($N=226$). Trait abbreviations:
O = Openness, C = Conscientiousness, E = Extroversion,
A = Agreeableness, and N = Neuroticism.}
\label{tab:intro-post-class-distribution}
\end{table}

\begin{table}[!t]
\centering
\small
\begin{tabular*}{\columnwidth}{
    @{\extracolsep{\fill}}lcr@{}
}
\toprule
\textbf{Trait} & \textbf{Value} & \textbf{Proportion} \\
\midrule
\textbf{cOPN}
    & 1 & 0.548780 (54.88\%) \\
    & 0 & 0.451220 (45.12\%) \\
\addlinespace

\textbf{cCON}
    & 1 & 0.542683 (54.27\%) \\
    & 0 & 0.457317 (45.73\%) \\
\addlinespace

\textbf{cEXT}
    & 1 & 0.518293 (51.83\%) \\
    & 0 & 0.481707 (48.17\%) \\
\addlinespace

\textbf{cAGR}
    & 1 & 0.548780 (54.88\%) \\
    & 0 & 0.451220 (45.12\%) \\
\addlinespace

\textbf{cNEU}
    & 1 & 0.451220 (45.12\%) \\
    & 0 & 0.548780 (54.88\%) \\
\bottomrule
\end{tabular*}
\caption{Binary division of classification buckets for BFI traits
from the myPersonality single-post-per-user subset ($N=164$). Trait abbreviations
follow those used by the underlying myPersonality dataset:
cOPN = Openness, cCON = Conscientiousness, cEXT = Extroversion,
cAGR = Agreeableness, and cNEU = Neuroticism.}
\label{tab:facebook_distro}
\end{table}

\begin{table}[!t]
\centering
\small
\begin{tabular*}{\columnwidth}{
    @{\extracolsep{\fill}}lcr@{}
}
\toprule
\textbf{Trait} & \textbf{Value} & \textbf{Proportion} \\
\midrule
\textbf{cOPN}
    & 1 & 0.515397 (51.54\%) \\
    & 0 & 0.484603 (48.46\%) \\
\addlinespace

\textbf{cCON}
    & 1 & 0.508104 (50.81\%) \\
    & 0 & 0.491896 (49.19\%) \\
\addlinespace

\textbf{cEXT}
    & 1 & 0.517423 (51.74\%) \\
    & 0 & 0.482577 (48.26\%) \\
\addlinespace

\textbf{cAGR}
    & 1 & 0.530794 (53.08\%) \\
    & 0 & 0.469206 (46.92\%) \\
\addlinespace

\textbf{cNEU}
    & 1 & 0.499595 (49.96\%) \\
    & 0 & 0.500405 (50.04\%) \\
\bottomrule
\end{tabular*}
\caption{Distribution of binary Big Five personality labels in the
full, original Pennebaker-King Essays dataset ($N=2468$). Trait abbreviations
follow those used by the underlying Pennebaker-King Essays dataset:
cOPN = Openness, cCON = Conscientiousness, cEXT = Extroversion,
cAGR = Agreeableness, and cNEU = Neuroticism.}
\label{tab:essays-class-distribution}
\end{table}

\begin{table}[t]
\centering
\small
\begin{tabular*}{\columnwidth}{
    @{\extracolsep{\fill}}lcr@{}
}
\toprule
\textbf{Trait} & \textbf{Value} & \textbf{Proportion} \\
\midrule
\textbf{cOPN}
    & 1 & 0.513274 (51.33\%) \\
    & 0 & 0.486726 (48.67\%) \\
\addlinespace

\textbf{cCON}
    & 1 & 0.508850 (50.88\%) \\
    & 0 & 0.491150 (49.12\%) \\
\addlinespace

\textbf{cEXT}
    & 1 & 0.517699 (51.77\%) \\
    & 0 & 0.482301 (48.23\%) \\
\addlinespace

\textbf{cAGR}
    & 1 & 0.535398 (53.54\%) \\
    & 0 & 0.464602 (46.46\%) \\
\addlinespace

\textbf{cNEU}
    & 1 & 0.500000 (50.00\%) \\
    & 0 & 0.500000 (50.00\%) \\
\bottomrule
\end{tabular*}
\caption{Distribution of binary Big Five personality labels in the
smaller Pennebaker-King Essays subset ($N=226$, $seed=67$).
Trait abbreviations: cOPN = Openness, cCON = Conscientiousness,
cEXT = Extroversion, cAGR = Agreeableness, and cNEU = Neuroticism.}
\label{tab:essays-subset}
\end{table}

\subsection{Lexicon Categories for Ablation}
This section shows the full list of retained lexicon subcategories from Empath and NRC, shown in Table~\ref{tab:lexicon}

\subsection{Complete Results}
The full statistical output of the described experiments are included here. 

\begin{table}[!t]
\centering
{\small
\begin{tabularx}{\columnwidth}{
    @{}
    >{\raggedright\arraybackslash}p{0.16\columnwidth}
    >{\raggedright\arraybackslash}p{0.23\columnwidth}
    >{\raggedright\arraybackslash}X
    @{}
}
\toprule
\textbf{Lexicon source}
& \textbf{Category group}
& \textbf{Retained subcategories} \\
\midrule

Empath
& Positive affect
& positive\_emotion, joy, cheerfulness, contentment, affection,
love, optimism, pride, zest \\
\midrule

Empath
& Negative affect
& negative\_emotion, sadness, disappointment, suffering, torment,
nervousness, fear, timidity, anger, rage, aggression, irritability,
exasperation, hate, disgust, shame \\
\midrule

Empath
& Social-affective stance
& sympathy, emotional, warmth \\
\midrule

Empath
& Social/normative language
& swearing\_terms \\
\midrule

Empath
& Cognitive style
& thinking, order, confusion, anticipation, deception \\
\midrule

NRC
& Core emotions and polarity
& joy, sadness, anger, fear, disgust, surprise, positive, negative \\

\bottomrule
\end{tabularx}
}
\caption{Psycholinguistic and affective lexicon categories retained during ablation.}
\label{tab:lexicon}
\end{table}

\subsubsection{Examples of Pre-Ablated and Post-Ablated Text}
Below is a contiguous essay excerpt before and after content masking. The model received the complete essay in each condition; only the corresponding excerpt is shown. Each underscore represents one or more consecutive masked tokens.

\begin{quote}
\textbf{Original Essay Excerpt:}
I'm not saying that I'm perfect. but I've learned over the past few years about what I want out of life and what I don't want. I'm living my life the way I want to. as stress-free as possible and as happy as possible. When I'm put into these stupid situations it just makes life that much harder and it sucks!

\textbf{Ablated:}
I'm not \_ that I'm perfect. but I've learned over the \_ few \_ about what I \_ out of \_ and what I don't \_ . I'm \_ my \_ the \_ I \_ to. \_ stress-\_ as \_ and \_ happy as \_ . When I'm \_ into these stupid \_ it just \_ much harder and it sucks!
\end{quote}

\subsection{Prompts}

Listing 1 shows the required format JSON example.
Listing 2 shows the tool call definition that was then used
in the model API call to return model outputs in the required structured format. The classification field was
required in every condition. The justification field
was required for the \textit{Basic+Ex} and \textit{Linguistic+Ex} conditions
and omitted for the \textit{Basic} condition.

\begin{lstlisting}[
  float=!t,
  captionpos=t,
  basicstyle=\ttfamily\footnotesize,
  frame=tb,
  breaklines=true,
  columns=fullflexible,
  keepspaces=true,
  showstringspaces=false,
  numbers=left,
  numberstyle=\tiny,
  caption={Unified JSON response format. The justification field is included only in explanation-required conditions.},
  label={lst:json-unified}
]
expected_format = """
{
  "Openness": {
    "classification": "low/high"
  },
  "Conscientiousness": {
    "classification": "low/high"
  },
  "Extroversion": {
    "classification": "low/high"
  },
  "Agreeableness": {
    "classification": "low/high"
  },
  "Neuroticism": {
    "classification": "low/high"
  }
}
"""
\end{lstlisting}

\begin{lstlisting}[
  float=!t,
  captionpos=t,
  basicstyle=\ttfamily\small,
  frame=tb,
  breaklines=true,
  columns=fullflexible,
  keepspaces=true,
  showstringspaces=false,
  numbers=left,
  numberstyle=\tiny,
  caption={Abridged structured tool schema used in the Basic+Ex and Linguistic+Ex conditions. The remaining four traits used the same object structure as Openness.},
  label={lst:tool-schema-explanation}
]
tools = [
{
  "type": "function",
  "function": {
    "name": "record_personality_profile",
    "description": "Records the Big Five personality traits and a justification for each classification.",
    "parameters": {
      "type": "object",
      "properties": {
        "Openness": {
          "type": "object",
          "description": "The analysis for the Openness trait.",
          "properties": {
            "classification": {
              "type": "string",
              "enum": ["low", "high"]
            },
            "justification": {
              "type": "string",
              "description": "Reasoning based on a direct quote from the text."
            }
          },
          "required": [
            "classification",
            "justification"
          ]
        }

        # Structure repeats for the other traits
      },
      "required": [
        "Openness",
        "Conscientiousness",
        "Extroversion",
        "Agreeableness",
        "Neuroticism"
      ]
    }
  }
}
]
\end{lstlisting}

Table~\ref{tab:prompt_conditions} shows the full prompts, system and user, used in the
study.

\begin{table}[!t]
\centering
\small
\begin{tabularx}{\columnwidth}{
    @{}
    >{\raggedright\arraybackslash}p{0.23\columnwidth}
    >{\raggedright\arraybackslash}X
    @{}
}
\hline
\textbf{Condition} & \textbf{Basic} \\
\hline
\textbf{System message} &
You are an expert in inferring Big-5 personality traits from text. \\
\textbf{User message} &
Analyze this text for personality traits. Classify each Big Five trait as either 'low' or 'high'. Return your results in the following JSON format without explanation: \\
\hline

\textbf{Condition} & \textbf{Basic+Ex} \\
\hline
\textbf{System message} &
You are an expert in inferring Big-5 personality traits from text. Your justifications are always evidence-base and grounded in direct quotes from the text. \\
\textbf{User message} &
Your task is to analyze the provided text, classify the author's Big Five personality traits, and provide a detailed justification for your analysis. Classify each Big Five trait as either 'low' or 'high', no deviation allowed. Return your detailed justification and results in the following JSON format: \\
\hline

\textbf{Condition} & \textbf{Linguistic+Ex} \\
\hline
\textbf{System message} &
You are an expert in inferring Big-5 personality traits from linguistic patterns in text. \\
\textbf{User message} &
Analyze this text for personality traits based only on writing style, word choice, and language patterns. Classify each Big Five trait as either 'low' or 'high'. If the linguistic evidence is insufficient, make your best inference based on available patterns, but always provide a classification, and note this in your explanation. Return your explanation and results in the following JSON format: \\
\hline
\end{tabularx}
\caption{Prompt conditions used for zero-shot Big Five prediction. Prompts are reproduced as used. Each user message was followed by an expected JSON-format example and the target text sample. Outputs were collected through required structured tool calls.}
\label{tab:prompt_conditions}
\end{table}

\subsubsection{Model Self-Explanation Results}
Below is an example of an original student introduction text, the model's prediction, and the explanation for its reasoning the model provided. Incidentally, this was a correct prediction, though the self-explanation listed topical content. This is one example of an explanation tagged 'DEM'.

\begin{quote}
\textbf{Introduction post:}

``Hello!

My name is [NAME], and I currently live in [LOCATION].
This is my first [PROGRAM] course!

I currently work as a software engineer at [ORGANIZATION].
Outside of work, I like to play sports and hang out with
friends/family.

From this class, I hope to develop a strong foundation in [FIELD].

One interesting fact about me is that I also have a
[SECOND OCCUPATION].``

\textbf{Prediction:} High Extraversion

\textbf{Explanation:}
``The text expresses enjoyment of social activities, such as playing
sports and spending time with friends and family, indicating
sociability.``
\end{quote}

Below is another example, this time tagged as 'LIN'. To preserve anonymous review, identifying spans in the displayed example were replaced with placeholders. The model received the complete original post. Interestingly, the model's prediction was correct in this case, as well.

\begin{quote}
\textbf{Introduction post:}

``Hello,

I am [NAME]. Currently, I am living in [LOCATION], and I am a
[OCCUPATION] by day and a [DEGREE PROGRAM] student by night. I am
specifically interested in this course, as I would like to learn more
about [FIELD] and determine how to implement it in my current research.
I took [COURSE] and [COURSE] last semester. That was a fun,
occasionally stressful, and excellent learning experience. I am
looking forward to this semester, and I hope to learn a lot. :)``

\textbf{Prediction:} Low Extraversion

\textbf{Explanation:}
``The text does not include social or outgoing language; it focuses more
on personal academic interests rather than social interactions.``
\end{quote}

Table~\ref{tab:XAI} shows the counts for text content broad category markers for correct/incorrect and across total predictions. Additionally, it shows the confusion matrix of cue counts for text content.

\begin{table*}[!t]
\centering
\small
\setlength{\tabcolsep}{2pt}

\begin{tabular}{@{}
p{0.055\textwidth}
p{0.095\textwidth}
p{0.095\textwidth}
p{0.095\textwidth}
p{0.145\textwidth}
p{0.145\textwidth}
p{0.145\textwidth}
p{0.145\textwidth}
@{}}

\hline\hline

\textbf{Prompt} &
\multicolumn{3}{c}{\textbf{Explanation-tag counts}} &
\multicolumn{4}{c}{\textbf{Topical cues by prediction outcome}} \\

\cline{2-4}
\cline{5-8}

&
\textbf{Correct} &
\textbf{Incorrect} &
\textbf{Total} &
\textbf{Correct high} &
\textbf{Correct low} &
\textbf{Incorrect high} &
\textbf{Incorrect low} \\

\hline

Basic &
LIN 68; DEM 48; CMB 19 &
LIN 45; DEM 29; CMB 17 &
LIN 113; DEM 77; CMB 36 &
\textbf{S/F 14}; family (3), board games (2), friends (2),
band (2), concerts (2), dog (2) &
\textbf{S/F 3}; kids (1), Netflix (1), coding (1), gaming (1) &
\textbf{S/F 14}; friends (4), gaming (4), concerts (3) &
\textbf{S/F 10}; family (3), traveling (2) \\

\hline

Linguistic &
LIN 87; DEM 29; CMB 7; OTH 1 &
LIN 66; DEM 25; CMB 11; OTH 0 &
LIN 153; DEM 54; CMB 18; OTH 1 &
\textbf{S/F 34}; family (4), band (3), board games (3),
concerts (3), friends (3), dog (2) &
\textbf{S/F 19}; reading (7), gaming (5), chess (2),
dogs (2), Netflix (2) &
\textbf{S/F 19}; family (6), travel (5), friends (3),
concerts (3) &
\textbf{S/F 6}; gaming (3), board/card games (3) \\

\hline\hline
\end{tabular}
\caption{GPT-4o-mini Extroversion explanation audit for student
introduction posts. Tag columns report explanation-category counts;
topical-cue columns report sports/fitness and other cue frequencies.
LIN = linguistic; DEM = demographic/topical; CMB = combined;
OTH = indeterminate.}
\label{tab:XAI}
\end{table*}

\subsubsection{Full Prediction and Ablation Results}
This section presents the remaining full Spearman r/ cohen's k results across all models that were not included in the main body, for Posts, Essays, and Facebook. Only Ablation results for GPT-4o were included as other models either failed to output results reaching significance or failed to complete the task consistently at all.

\subsubsection{Pre-Ablated Predictions from Text}
Table~\ref{tab:original} shows the full results on the non-ablated text, for all 3 datasets.

\begin{table*}[!t]\centering\small\footnotesize\setlength{\tabcolsep}{5pt}
\begin{tabular}{llccccc}
\hline\hline
Prompt & Model & O & C & E & A & N \\
\hline
\multicolumn{7}{l}{\textit{Posts ($n{=}226$)}} \\
\hline
\textbf{BASIC} & GPT-4o & -.039/-.024 & .055/.027 & .101/.088 & .101/.074 & -.058/-.009 \\
 & GPT-4o-mini & -.032/-.017 & .055/.027 & .209$^{**}$/.153$^{**}$ & .105/.049 & .077/.012 \\
 & 5.4-mini & -.022/-.009 & .033/.019 & .162$^{*}$/.151$^{*}$ & .105/.049 & -.082/-.018 \\
 & o3-mini & -.032/-.017 & -.058/-.033 & .211$^{**}$/.178$^{**}$ & -.026/-.009 & -??? \\
 & Claude Haiku & -.022/-.009 & -.041/-.017 & .238$^{***}$/.165$^{***}$ & -??? & -??? \\
\textbf{LINEX} & GPT-4o & -.056/-.044 & .033/.019 & .091/.088 & .110/.094 & -??? \\
 & GPT-4o-mini & -.045/-.031 & -.026/-.023 & .226$^{***}$/.173$^{**}$ & -.036/-.017 & .024/.009 \\
 & 5.4-mini & -??? & -.058/-.033 & .138$^{*}$/.137 & -.063/-.046 & -.022/-.006 \\
 & o3-mini & .066/.045 & -.058/-.033 & .101/.088 & -.058/-.039 & -??? \\
 & Claude Haiku & -.022/-.009 & .033/.019 & .205$^{**}$/.136$^{**}$ & -??? & -??? \\
\hline
\multicolumn{7}{l}{\textit{Essays ($n{=}226$)}} \\
\hline
\textbf{BASIC} & GPT-4o & .221$^{***}$/.172$^{**}$ & .139$^{*}$/.122 & .102/.097 & .200$^{**}$/.195$^{**}$ & .133$^{*}$/.062 \\
 & GPT-4o-mini & .209$^{**}$/.142$^{**}$ & .115/.097 & .101/.100 & .144$^{*}$/.137$^{*}$ & .198$^{**}$/.142$^{**}$ \\
 & 5.4-mini & .263$^{***}$/.198$^{***}$ & .149$^{*}$/.149$^{*}$ & .151$^{*}$/.150$^{*}$ & .193$^{**}$/.178$^{**}$ & .102/.062 \\
 & o3-mini & .269$^{***}$/.178$^{***}$ & .126/.106 & .111/.106 & .128/.125 & .168$^{*}$/.097$^{*}$ \\
 & Claude Haiku & .132$^{*}$/.047 & .114/.104 & .183$^{**}$/.183$^{**}$ & .165$^{*}$/.144$^{*}$ & .124/.062 \\
\textbf{LINEX} & GPT-4o & .215$^{**}$/.162$^{**}$ & .150$^{*}$/.131$^{*}$ & .180$^{**}$/.179$^{*}$ & .198$^{**}$/.198$^{**}$ & .119/.071 \\
 & GPT-4o-mini & .204$^{**}$/.133$^{**}$ & .087/.071 & .158$^{*}$/.157$^{*}$ & .185$^{**}$/.164$^{**}$ & .175$^{**}$/.097$^{*}$ \\
 & 5.4-mini & .263$^{***}$/.198$^{***}$ & .115/.114 & .151$^{*}$/.151$^{*}$ & .165$^{*}$/.145$^{*}$ & .133$^{*}$/.062 \\
 & o3-mini & .271$^{***}$/.188$^{***}$ & .152$^{*}$/.131$^{*}$ & .126/.122 & .109/.106 & .166$^{*}$/.106$^{*}$ \\
 & Claude Haiku & .180$^{**}$/.094$^{*}$ & .126/.113 & .147$^{*}$/.141$^{*}$ & .185$^{**}$/.165$^{**}$ & .089/.044 \\
\hline
\multicolumn{7}{l}{\textit{Facebook (per-user, $n{=}164$)}} \\
\hline
\textbf{BASIC} & GPT-4o & -.020/-.020 & .163$^{*}$/.148 & .064/.063 & -.016/-.016 & .184$^{*}$/.183$^{*}$ \\
 & GPT-4o-mini & -.015/-.015 & .087/.051 & .088/.088 & .078/.078 & .081/.078 \\
 & 5.4-mini & -.025/-.024 & .071/.061 & .024/.024 & .046/.046 & .076/.076 \\
 & o3-mini & -.007/-.007 & .078/.064 & -.000/-.000 & .049/.048 & .117/.110 \\
 & Claude Haiku & -.060/-.056 & .127/.119 & .015/.014 & -.042/-.040 & .094/.094 \\
\textbf{LINEX} & GPT-4o & .053/.051 & .118/.099 & .060/.060 & .037/.037 & .166$^{*}$/.164 \\
 & GPT-4o-mini & .056/.056 & .071/.041 & -.022/-.022 & .116/.115 & .067/.064 \\
 & 5.4-mini & ??? & .122/.106 & -.038/-.038 & -.010/-.009 & .083/.082 \\
 & o3-mini & .051/.051 & .062/.052 & .041/.040 & -.088/-.085 & .098/.096 \\
 & Claude Haiku & .004/.004 & .167$^{*}$/.149$^{*}$ & .080/.064 & -.058/-.053 & .080/.080 \\
\hline\hline
\end{tabular}
\caption{Cells: Spearman $\rho$/Cohen's $\kappa$. Stars on Spearman $p$, on $\kappa$ from $\chi^2$ $p$: $^{*}p{<}.05$, $^{**}p{<}.01$, $^{***}p{<}.001$. O=Openness, C=Conscientiousness, E=Extroversion, A=Agreeableness, N=Neuroticism. Facebook ablation run for GPT-4o only. Values of ??? indicate single class prediction only.}
\label{tab:original}
\end{table*}

\subsubsection{Post-Ablated Predictions from Text}
Table~\ref{tab:ablated} shows the final full results on the ablated text, for all 3 datasets.

\begin{table*}[!t]\centering\small\footnotesize\setlength{\tabcolsep}{5pt}
\begin{tabular}{llccccc}
\hline\hline
Prompt & Model & O & C & E & A & N \\
\hline
\multicolumn{7}{l}{\textit{Posts ($n{=}226$)}} \\
\hline
\textbf{BASIC} & GPT-4o & -.001/-.001 & -.050/-.048 & .017/.006 & -??? & .052/.023 \\
 & GPT-4o-mini & .015/.013 & -.045/-.044 & .103/.038 & -??? & .031/.014 \\
 & 5.4-mini & .049/.036 & -.048/-.048 & .153$^{*}$/.139$^{*}$ & .070/.056 & .024/.009 \\
 & o3-mini & .145$^{*}$/.142 & -.118/-.097 & .139$^{*}$/.114 & .082/.082 & .052/.023 \\
 & Claude Haiku & -.026/-.025 & -.057/-.057 & .061/.033 & -.036/-.017 & .027/.012 \\
\textbf{LINEX} & GPT-4o & -.045/-.045 & -.004/-.004 & .098/.097 & .072/.072 & .092/.062 \\
 & GPT-4o-mini & -.035/-.032 & -.111/-.068 & .062/.058 & .045/.044 & .078/.054 \\
 & 5.4-mini & -.008/-.008 & -.050/-.048 & .075/.029 & -??? & .027/.012 \\
 & o3-mini & -.038/-.038 & -.001/-.001 & .110/.090 & .059/.059 & .051/.020 \\
 & Claude Haiku & -.026/-.025 & -.057/-.057 & .061/.033 & -.036/-.017 & .027/.012 \\
\hline
\multicolumn{7}{l}{\textit{Essays ($n{=}226$)}} \\
\hline
\textbf{BASIC} & GPT-4o & .224$^{***}$/.215$^{**}$ & .086/.057 & .076/.057 & .073/.053 & .030/.009 \\
 & GPT-4o-mini & .160$^{*}$/.139$^{*}$ & .038/.023 & .064/.062 & .148$^{*}$/.148$^{*}$ & .144$^{*}$/.053 \\
 & 5.4-mini & .175$^{**}$/.146$^{*}$ & .078/.077 & .091/.088 & .182$^{**}$/.176$^{**}$ & .137$^{*}$/.080 \\
 & o3-mini & .171$^{**}$/.115$^{*}$ & .045/.024 & .033/.027 & .112/.094 & .110/.035 \\
 & Claude Haiku & .180$^{**}$/.094$^{*}$ & .084/.071 & .152$^{*}$/.152$^{*}$ & .247$^{***}$/.241$^{***}$ & .116/.027 \\
\textbf{LINEX} & GPT-4o & .216$^{**}$/.183$^{**}$ & .139$^{*}$/.092 & .194$^{**}$/.172$^{**}$ & .230$^{***}$/.223$^{***}$ & .128/.044 \\
 & GPT-4o-mini & .245$^{***}$/.232$^{***}$ & .106/.050 & .069/.064 & .096/.095 & .144$^{*}$/.053 \\
 & 5.4-mini & .161$^{*}$/.151$^{*}$ & .087/.085 & .144$^{*}$/.139$^{*}$ & .231$^{***}$/.215$^{***}$ & ??? \\
 & o3-mini & .128/.100 & .071/.040 & .119/.097 & .132$^{*}$/.111 & .096/.035 \\
 & Claude Haiku & .192$^{**}$/.104$^{**}$ & .091/.072 & .110/.102 & .220$^{***}$/.214$^{**}$ & .134$^{*}$/.035 \\
\hline
\multicolumn{7}{l}{\textit{Facebook (per-user, $n{=}164$)}} \\
\hline
\textbf{BASIC} & GPT-4o & .066/.052 & .088/.044 & .028/.025 & -.027/-.022 & -.058/-.056 \\
\textbf{LINEX} & GPT-4o & .104/.083 & .105/.055 & .108/.098 & .004/.004 & .062/.062 \\
\hline\hline
\end{tabular}
\caption{Cells: Spearman $\rho$/Cohen's $\kappa$. Stars on Spearman $p$, on $\kappa$ from $\chi^2$ $p$: $^{*}p{<}.05$, $^{**}p{<}.01$, $^{***}p{<}.001$. O=Openness, C=Conscientiousness, E=Extroversion, A=Agreeableness, N=Neuroticism. Facebook ablation run for GPT-4o only. Values of ??? indicate single class prediction only.}
\label{tab:ablated}
\end{table*}

\end{document}